\documentclass{article}
\usepackage{arxiv}

\usepackage[utf8]{inputenc} 
\usepackage[T1]{fontenc}    
\usepackage{hyperref}       
\usepackage{url}            
\usepackage{booktabs}       
\usepackage{amsfonts}       
\usepackage{nicefrac}       
\usepackage{microtype}      
\usepackage{graphicx}
\usepackage{natbib}
\usepackage{doi}

\usepackage{microtype}
\usepackage{subcaption}
\usepackage{wrapfig}
\usepackage{hyperref}
\usepackage{amsmath}
\usepackage{amssymb}
\usepackage{mathtools}
\usepackage{amsthm}
\usepackage{xcolor}
\usepackage[most]{tcolorbox}

\usepackage{multirow}
\usepackage{bm} 
\usepackage{paralist}
\usepackage{multicol}
\usepackage{url}

\usepackage[ruled,noresetcount,vlined,linesnumbered]{algorithm2e}
\SetKwInput{KwInput}{Input}                
\SetKwInput{KwOutput}{Output}

\DeclareMathOperator*{\argmax}{argmax}
\DeclareMathOperator*{\argmin}{argmin}

 \newcommand{\cB}{\mathcal{B}} 
\newcommand{\cC}{\mathcal{C}} 
\newcommand{\cE}{\mathcal{E}}

\newcommand{\cM}{\mathcal{M}} 
\newcommand{\cO}{\mathcal{O}} \newcommand{\cP}{\mathcal{P}}

\newcommand{\EE}{\mathbb{E}} \newcommand{\RR}{\mathbb{R}}

\newcommand\DCG{\textsl{DCG}}

\DeclareMathOperator{\proj}{proj}

\usepackage{esvect}

\usepackage{letltxmacro}
\LetLtxMacro\orgvdots\vdots
\LetLtxMacro\orgddots\ddots

\title{Learning Fair Ranking Policies via Differentiable Optimization of Ordered Weighted Averages}

\renewcommand{\shorttitle}{Learning Fair Ranking Policies via Differentiable Optimization of OWAs}

\date{}

\author{ 
\href{https://orcid.org/0000-0002-6367-9626}{\includegraphics[scale=0.06]{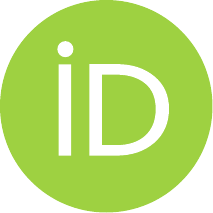}\hspace{1mm}My H.~Dinh}\thanks{equal contribution} \\
	University of Virginia\\
	Charlottesville, VA, USA\\
	\texttt{fqw2tz@virginia.edu}\\
	\And
	\href{https://orcid.org/0000-0002-4499-8066}
    {\includegraphics[scale=0.06]{orcid.pdf}\hspace{1mm}
    James Kotary}$^*$ \\
	University of Virginia\\
	Charlottesville, VA, USA\\
	\texttt{jk4pn@virginia.edu} \\
    \And
     \href{https://orcid.org/0000-0002-1381-6776}{\includegraphics[scale=0.06]{orcid.pdf}\hspace{1mm}Ferdinando Fioretto
     } \\
	University of Virginia\\
	Charlottesville, VA, USA\\
	\texttt{fioretto@virginia.edu} \\
}

\begin{document}
\maketitle\allowdisplaybreaks\sloppy

\begin{abstract}
Learning to Rank (LTR) is one of the most widely used machine learning applications. It is a key component in platforms with profound societal impacts, including job search, healthcare information retrieval, and social media content feeds. Conventional LTR models have been shown to produce biases results, 
stimulating a discourse on how to address the disparities introduced by ranking systems that solely prioritize user relevance.  
However, while several models of fair learning to rank have been proposed, they suffer from deficiencies either in accuracy or efficiency, thus limiting their applicability to real-world ranking platforms. 
This paper shows how efficiently-solvable fair ranking models, based on the optimization of Ordered Weighted Average (OWA) functions, can be integrated into the training loop of an LTR model to achieve favorable balances between fairness, user utility, and runtime efficiency. In particular, this paper is the first to show how to backpropagate through constrained optimizations of OWA objectives, enabling their use in integrated prediction and decision models. 
\end{abstract}

\section{Introduction}
\label{sec:intro}


Ranking models have become a pervasive aspect of everyday life. They are at the center of how people find information online, serving as the main mechanisms by which we interact with products, content, and other people. In these systems, the items to be ranked are videos, job candidates, research papers, and almost anything else. As models based on machine learning, they are primarily trained to provide maximum utility to users, by serving the results deemed most relevant to their search queries. In the modern economy of information, the position of an item in the ranking has a strong influence on its exposure, selection, and, ultimately its economic success. 

Because of this influence, increasing attention has been placed on the disparate impacts of ranking systems on underrepresented groups. In these data-driven systems, the relevance of an item is measured by implicit feedback from users such as clicks and dwell times. As such, the disparate impacts of rankings can go well beyond their immediate effects. Disproportionate exposure in rankings results leads to higher selection rates, in turn boosting relevance scores based on implicit feedback \citep{yadav2019fair,sun2020evolution}. This can create self-reinforcing feedback loops, leading to winner-take-all dynamics. The ability to control these disparate impacts is essential in order to avoid reinforcement of systemic biases, ensure the health and stability of online markets, and implement anti-discrimination measures \citep{edelman2017racial, singh2019policy}. 

When search results are ranked purely based on relevance, disparate exposure between groups may be greatly increased in order to achieve marginal gains in relevance. For example, in a job search system it is possible for male candidates to receive overwhelmingly more exposure even when female candidates may have been rated only marginally lower in relevance.
It has indeed been found in  \citep{elbassuoni2019exploring} that in a job candidate ranking system, small differences in relevance can lead to large differences in exposure for candidates from a minority group. Thus, fairness-aware ranking models often suffer little to no degradation to user utility when compared to their conventional counterparts \citep{zehlike2020reducing}.

On the other hand, these fair learning to rank models are more difficult to design, since they require the outputs of a machine learning model to obey potentially complex constraints while simultaneously achieving high relevance. For this reason, conventional learning to rank methods are often maladjusted to incorporate fairness. For example, the popular listwise learning to rank method, through modifications to its loss function, is only capable of modeling fairness of exposure in the top ranking position \citep{zehlike2020reducing}. 
In an alternative paradigm, first investigated by \cite{kotary2022end}, a fair ranking linear programming model is integrated with LTR in an end-to-end training process. By incorporating fair ranking optimization into the model's training loop, rather than in post-processing, utility of the downstream fair ranking policies can be maximized as a loss function. This allows 
to provide fairness guarantees at the level of each predicted ranking policy, and precise control over the fairness-utility trade-off. However, the method comes with a significant computational cost, as it requires solving a large optimization problem for each sample in each training iteration, challenging its application to real-world ranking systems. A further limitation of fair LTR systems, including that of \citep{kotary2022end}, is their inability to effectively deal with multi-group fairness criteria (i.e., going beyond binary group treatment), which are overwhelmingly common in real world applications. 

\begin{figure}
    \centering
    \includegraphics[width=1\textwidth]{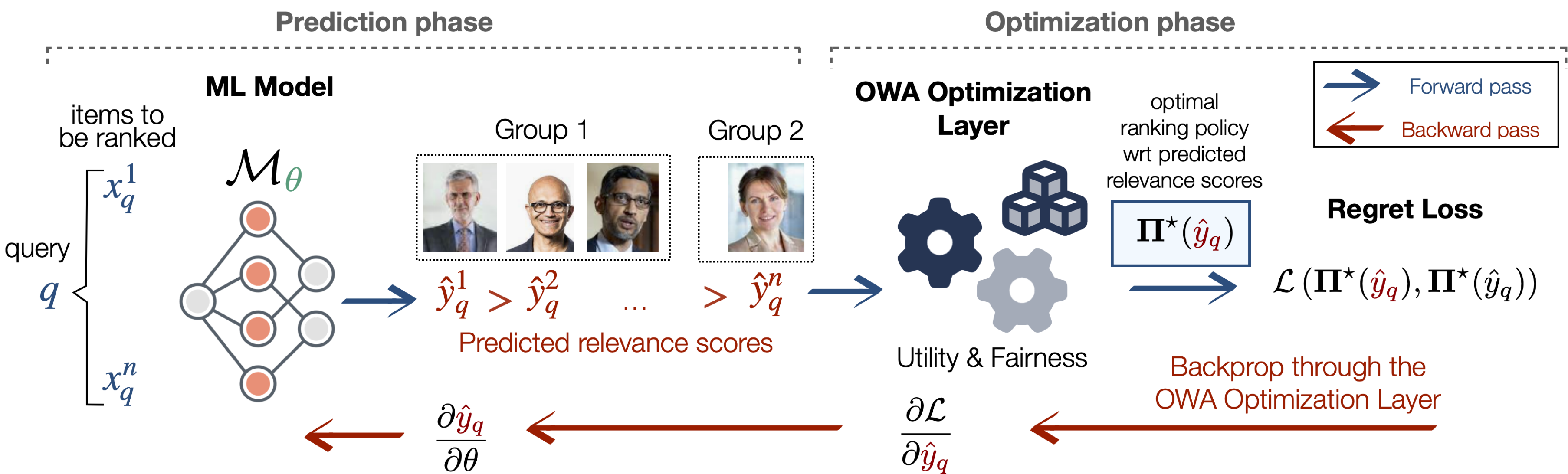}
    \caption{The differentiable optimization module proposed in SOFaiR. Its forward pass is calculated by an efficient Frank-Wolfe method, and its backward pass computes the SPO+ subgradient of the OWA problem's regret due to prediction error.}
    \label{fig:pipeline}
\end{figure}

{\bf Contributions.} To address these limitations, this paper makes the following novel contributions:
\textbf{(1)} It shows how to adopt an alternative approach based on Ordered Weighted Averages (OWA) to design efficient policy optimization modules for the fair learning to rank setting. \textbf{(2)} For the first time, it shows how to backpropagate gradients through the highly discontinuous optimization of OWA functions, enabling its use in end-to-end learning. \textbf{(3)} The resulting end-to-end optimization and learning scheme, called Smart OWA Optimization for Fair Learning to Rank ({\bf SOFaiR}), is compared with contemporary fair LTR methods, demonstrating not only substantial advantages in fairness over previous fair LTR, but also advantages in efficiency and modeling flexibility over the end-to-end fair LTR scheme of \citep{kotary2022end}. A schematic illustration of the proposed scheme is depicted in Figure \ref{fig:pipeline}. 

These contributions are significant: They demonstrate that by incorporating modern fair ranking optimization techniques, the integration of post-processing optimization models in end-to-end LTR training can be a viable and scalable paradigm to achieve highly accurate learning to rank system that also provides strong fairness properties. 


\section{Preliminaries}
\label{sec:preliminaries}

Throughout the paper, 
vectors and matrices are denoted in bold font. The inner product of two vectors $\mathbf{a}$ and $\mathbf{b}$ is written $\mathbf{a}^T \mathbf{b}$, while the outer product is $\mathbf{a} \; \mathbf{b}^T$. For a matrix $\mathbf{M}$, the vector $\overrightarrow{\mathbf{M}}$ is formed by concatenation of its rows. A hatted vector $\hat{\mathbf{a}}$ is the prediction of a machine learning model, and a starred vector $\mathbf{a}^{\star}$ is the optimal solution to some optimization problem. The list of integers $\{1 \ldots n\}$ is written $\left[ n \right]$. When $\mathbf{a} \in \mathbb{R}^n$ and $\sigma$ is a permutation of $\left[ n \right]$, $\mathbf{a}_{\sigma}$ is the corresponding permuted vector. The vectors of all ones and zeros are denoted $\mathbf{1}$ and $\mathbf{0}$, respectively. Commonly used symbols throughout the paper are organized in Table \ref{table:symbols} for reference. 

\begin{table*}[!t]
    \centering
    \begin{minipage}{0.5\textwidth}
        \resizebox{\textwidth}{!}{
            \begin{tabular}{l l} 
                \toprule
                \textbf{Symbol} & \textbf{Semantic}\\
                \midrule
                $N$      & Size of the training dataset\\
                $n$      & Number of items to be ranked\\
                $m$      & Number of protected groups\\
                $\bm{x}_q = (x^i_q)_{i=1}^n$ & List of feature embeddings for items to rank, given query $q$\\
                $\bm{a}_q = (a^i_q)_{i=1}^n$ & Protected groups associated with items $x^i_q$\\
                $\bm{y}_q = (y^i_q)_{i=1}^n$ & Relevance scores for each of $n$ items given query $q$\\
                $G$ & The set of all protected group indicators\\
                $\cM_{\theta}$ & End-to-end trainable fair ranking model with weights $\theta$\\ 
                \\
                \bottomrule
            \end{tabular}
        }
    \end{minipage}%
    \begin{minipage}{0.5\textwidth}
        \resizebox{\textwidth}{!}{
            \begin{tabular}{l l}
                \toprule
                \textbf{Symbol} & \textbf{Semantic}\\
                \midrule
                $\sigma$ & A permutation of the list $[n]$ for some $n$\\
                $\mathbf{P}$ & A permutation matrix corresponding to some $\sigma$\\ 
                $\cP_n$ & The set of all permutations of $[n]$\\
                $\tau$ & The sorting operator\\
                $\Pi$ & A ranking policy, or its representative bistochastic matrix\\
                $\bm{u}(\Pi,y)$ & Expected utility of policy $\Pi$ under relevance scores $y$\\
                $\cB$ & Birkhoff Polytope, the convex set of all ranking policies\\
                $\cE(i, \sigma)$ & Exposure of item $i$\\[5pt]
                \bottomrule
            \end{tabular}
        }
    \end{minipage}
    \caption{Common symbols adopted throughout the paper.}
    \label{table:symbols}
\end{table*}

\subsection{Problem Setting and Goals}
Given a user query, the goal is to predict a ranking over $n$ items, in order of most to least relevant, with respect to the query. 
Relevance of each item to be ranked, with respect to a search query $q$, is generally measured by a vector of \emph{relevance scores} $\bm{y}_q \in \mathbb{R}^n$, often modeled on the basis of empirical observations such as historical click rates \citep{xu2010improving}. 
This setting considers a ground-truth dataset $(\bm{x}_q, \bm{a}_q, \bm{y}_q)_{q=1}^{N}$, where
 $\bm{x}_q \in \mathcal{X}$ is a list of feature vectors $(x_q^i)_{i=1}^n$, one for each of $n$ items to be ranked in response to query $q$.
$\bm{a}_q = (a_q^i)_{i=1}^n$ is a vector that indicates which (protected) group $g$ within domain $G$ to which each item belongs.
$\bm{y}_q = (y_q^i)_{i=1}^n \in \mathcal{Y}$  is a vector of relevance scores, for each item with respect to query $q$. 
For example, on an image web-search context as depicted in Figure \ref{fig:pipeline}, a query denotes the search keywords, e.g., ``CEO'', the vectors $x_q^i$ in $\bm{x}_q$ are feature embeddings for the images relative to $q$, each associated with a gender (attribute $a_q^i$), and the associated relevance scores $y_q^i$ describe the relevance of item $i$ to query $q$.

\noindent
Rankings can be viewed as \emph{permutations} which rearrange the order of a predefined \emph{item list}. Intermediate between the user input and final ranking is often a ranking \emph{policy} which produces discrete rankings (randomly or deterministically).

\smallskip\noindent{\bf Learning to Rank.} In {learning to rank (LTR)}, a ML model $\mathcal{M}_{\theta}$ is often adopted to estimate relevance scores $\hat{\mathbf{y}}_q$ of items given their features $\mathbf{x}_q$ relative to user query $q$ (see Figure \ref{fig:pipeline}). From this a ranking \emph{policy} $\mathbf{\Pi}$ is constructed. Its \emph{expected utility} $u$ is
\begin{equation}
\label{eq:utilty_def}
 u(\mathbf{\Pi}, \bf{y}_q) = \mathbb{E}_{\sigma \sim \mathbf{\Pi}} [\Delta(\sigma, \bf{y}_q)],
 \end{equation}
where $\mathbf{\Pi}$ is viewed as a distribution from which rankings $\sigma$ are sampled randomly, and their utility $\Delta$ is a measure of the overall relevance of a given ranking $\sigma$, with respect to given relevance scores $\textbf{y}_q$. Although its framework is applicable to any linear utility metric $\Delta$ for rankings, this paper uses the widely adopted Discounted Cumulative Gain (DCG):
\begin{equation}
    \Delta(\sigma, \bf{y}_q) = \DCG(\sigma, \bm{y}_q) = \sum_{i=1}^n y_q^i \bm{b}_{\sigma_i} = \mathbf{y}_q^T \mathbf{P}^{(\sigma)}  \bm{b},
\end{equation}
where $\mathbf{P}^{(\sigma)}$ is the corresponding permutation matrix, $\bm{y}_q$ are the true 
relevance scores, and $\bm{b}$ is a \emph{position bias} vector which models the probability that each position is viewed by a user, defined with elements $b_{j} = \nicefrac{{1}}{\log_2\left(1+j \right)}$, for $j \in [n]$.

\smallskip\noindent{\bf Ranking policy representation.} 
The methods of this paper adopt a particular representation of the ranking policy, as bistochastic matrix $\mathbf{\Pi} \in \RR^{n\times n}$, where $\mathbf{\Pi}_{jk}$ indicates the probability that item $j$ takes position $k$ in the ranking. The set of feasible ranking policies is expressed as $\mathbf{\Pi} \in \cB$ where $\cB$ is the \emph{Birkhoff Polytope}:
\begin{equation}
    \label{model:birkhoff_polytope}
    \cB = \{ \mathbf{\Pi} \;\; \textit{s.t.} \;\; \mathbf{1}^T \bm{\Pi} = \mathbf{1}, \;\;  \bm{\Pi} \, \mathbf{1} = \mathbf{1}, \;\; \mathbf{0} \leq \bm{\Pi}  \leq \mathbf{1} \}. 
\end{equation}
Its conditions on a matrix $\mathbf{\Pi}$ require, in the order of     \eqref{model:birkhoff_polytope}, that each column of $\mathbf{\Pi}$ sums to one, each row of $\mathbf{\Pi}$ sum to one, and each element of $\mathbf{\Pi}$ lie between $0$ and $1$. Each of these conditions is a linear constraint on the variables $\mathbf{\Pi}$.

Linearity of the DCG function \eqref{eq:utilty_def}  w.r.t. $\mathbf{P}$ allows it to commute with the expectation, leading to the practical closed form $ u(\mathbf{\Pi}, \bf{y}) =  \bm{y}^\top \mathbf{\Pi} \, \bm{b}$ for $u$ as a linear function of $\mathbf{\Pi}$:
\begin{equation}
    \label{eq:birkhoff_dcg_work}
        u(\mathbf{\Pi}, \bf{y}) = \mathbb{E}_{\sigma \sim \mathbf{\Pi}}  \Delta(\sigma, \bm{y})     = \mathbb{E}_{\sigma \sim \mathbf{\Pi}}  \, \left[ \bm{y}^\top \mathbf{P}^{(\sigma)} \, \bm{b} \right] = \bm{y}^\top \left(\mathbb{E}_{\sigma \sim \mathbf{\Pi}}  \,  \mathbf{P}^{(\sigma)} \right)\,\bm{b} 
          = \bm{y}^\top \mathbf{\Pi} \, \bm{b}.
\end{equation}
This is an important observation that enables the constrained optimization of utility functions on the policy $\bm{\Pi}$ in end-to-end differentiable pipelines, as discussed later in the paper. 

\subsection{Fairness of Exposure}
Item exposure is commonly adopted in ranking systems, where items in higher ranking positions receive more exposure, and it is with respect to this metric that fairness is concerned.  This paper aims at learning ranking policies that satisfy \textbf{group fairness of exposure}, while maintaining high relevance to user queries. The exposure $\cE (i, \sigma)$  of item $i$ within some ranking $\sigma$ is a function of only its position, with higher positions receiving more exposure than lower ones. Throughout the paper, the common modeling choice $\cE (i, \sigma) = b_{\sigma_i}$.

Notions of item exposure in rankings can also be extended to group exposure in ranking policies. The exposure of group $g$ in ranking $\sigma$ is measured by the mean exposure in $\sigma$ of items belonging to $g$. The exposure of group $g$ in ranking policy $\mathbf{\Pi}$ is the mean value of its exposure over all rankings sampled from the policy: 
\begin{equation}
    \label{eq:group_exposure}
    \cE_g(\mathbf{\Pi}) = \EE_{\substack{\sigma \sim \mathbf{\Pi} \\ i \sim [n]}} 
  \left[ \cE \left(i, \sigma \right) | a_q^i = g \right],
\end{equation}
and we let $\cE_G(\mathbf{\Pi})$ be the vector of values \eqref{eq:group_exposure} for each $g$ in $G$.
Derived similarly to \eqref{eq:birkhoff_dcg_work}, linearity of $\cE$ leads to a closed form for \eqref{eq:group_exposure} when $\mathbf{\Pi}$ is represented by a bistochastic matrix, where $\bm{1}_g$ indicates $1$ for items in $g$ and $0$ elsewhere \citep{singh2018fairness}:
\begin{equation}
  \label{eq:group_exposure_closed}
    \cE_g(\mathbf{\Pi}) = \bm{1}_g^T \mathbf{\Pi} \, \bm{b}.
\end{equation}

\noindent\textbf{Imposing fairness in LTR.} 
It is well-known that individual rankings $\sigma$, as discrete structures, cannot exactly satisfy most notions of individual or group fairness \citep{zehlike2017fa}. Therefore a common strategy in fair ranking optimization is to view ranking policies as random distributions of rankings, upon which a feasible notion of fairness can be imposed \emph{in expectation} \citep{zehlike2017fa,singh2018fairness,do2022optimizing}. 
For ranking policy $\mathbf{\Pi}$ and query $q$, fairness of exposure requires that every group indicated by $g \in G$
receives equal exposure on average over rankings produced by the policy. This condition can be expressed by requiring that the average exposure among items of each group is equal to the average exposure among all items:
\begin{equation}
\label{eq:fairness_of_exposure}
     \cE_g(\mathbf{\Pi}) = \cE_{\alpha}(\mathbf{\Pi}), \;\;\;\; \forall g \in G,
\end{equation}
where $\alpha$ is the group containing all items.
Enforcing the condition \eqref{eq:fairness_of_exposure} on each predicted policy $\mathbf{\Pi}$ is the mechanism by which protected groups are ensured equal exposure in SOFaiR. In the image search example, it corresponds to male and female candidates receiving equal exposure on average over rankings sampled from $\mathbf{\Pi}$.  The violation of fairness with respect to group $g$ is measured by the absolute gap in this condition:
\begin{equation}
\label{eq:fairness_violation}
  \nu_g(\mathbf{\Pi}) =\left| \; \cE_g(\mathbf{\Pi}) - \cE_{\alpha}(\mathbf{\Pi}) \; \right|.
\end{equation}
Note that group fairness encompasses individual item fairness is a special case, where each item belongs to a distinct group. While the fairness and utility metrics described above are the ones used throughout the paper, the methodology of the paper is compatible with any alternative metrics $u$ and $\cE$ which are \emph{linear} functions of the policy $\mathbf{\Pi}$. This is because the methodology of Sections \ref{sec:forward_pass} and \ref{sec:backward_pass} depend on linearity of \eqref{eq:birkhoff_dcg_work} and \eqref{eq:group_exposure_closed}.

\section{Limitations of Fair LTR Methods}
\label{sec:limitations}
Current Fair LTR models present a combination of the following limitations: {\bf (A)} inability to ensure fairness in each of its generated policies, {\bf (B)} inability or ineffectiveness to handle multiple protected groups, and {\bf (C)} inefficiency at training and inference time. This section reviews current fair LTR methods in light of these limiting factors. 

\begin{wrapfigure}[13]{r}{0.40\textwidth}
    \vspace{-12pt}
    \centering
    \includegraphics[width=\linewidth]{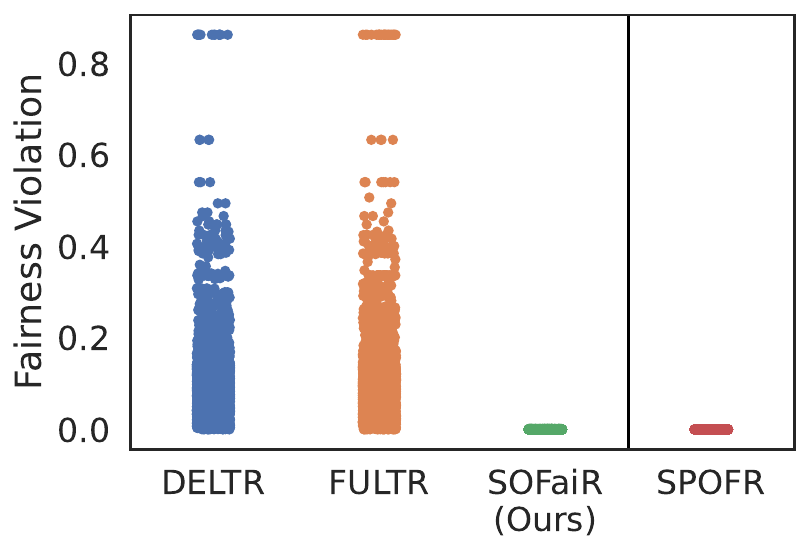}
    \vspace{-16pt}
        \caption{\small Yahoo-20: Fairness violation at query level. 
         }
    \label{fig:variance}
\end{wrapfigure}
\noindent
Regardless of how the policy is represented, fair learning to rank methods typically train a model $\mathcal{M}_{\theta}$ to find parameters $\theta^*$ that maximize its empirical utility, along with possibly a weighted penalty term $F$ which promotes fairness: 
\begin{equation}
    \label{eq:fair_ERM}
    \theta^*   = \argmax_{\bm{\theta}} \;\;
    \frac{1}{N} \sum_{q = 1}^N u(\mathcal{M}_{\bm{\theta}}(\bm{x}_q), \bm{y}_q) + \lambda \cdot F(\mathcal{M}_{\bm{\theta}}(\bm{x}_q))
\end{equation}
For example, the fair LTR method of \cite{zehlike2020reducing} (called DELTR) is based on listwise learning to rank \citep{cao2007learning}, and thus uses the model $\mathcal{M}_{\theta}$ to predict activation scores per each individual item, over which a softmax layer defines the probabilities of each item taking the top ranking position. Thus, \cite{zehlike2020reducing} can only use $F$ to encourage group fairness of exposure in the top position, leading to poor overall satisfaction of the fairness condition \eqref{eq:fairness_of_exposure} ({\bf limitation A}) as illustrated in Figure \ref{fig:variance}. 
To impose fairness over all ranking positions, \citep{singh2019policy} (FULTR) also uses softmax over the activations of $\mathcal{M}_{\theta}$ to define 
probabilities, which are sampled without replacement to generate rankings using a policy gradient method. However, this penalty-based method still does not ensure fairness in each predicted policy, as illustrated in Figure \ref{fig:variance}, since the penalty is imposed only \emph{on average} over all predicted policies (\textbf{limitation A}). 
By a similar reasoning, these methods do not translate naturally to the case of multigroup fairness, where $m>2$ (\textbf{limitation B}): Because the penalty $F$ must scalarize the collection of all group fairness violations \eqref{eq:fairness_violation} (by taking their overall sum), it is possible to reduce $F$ while increasing the exposure of a single outlier group \citep{kotary2022end}. 

Later work \citep{kotary2022end} shows how to overcome limitation A, 
by integrating the fair ranking optimization model of \cite{singh2018fairness} together with prediction of relevance scores $\hat{\mathbf{y}}_q = \cM_{\theta}$. 
The modeling of predicted policies $\bm{\Pi}$ as solutions to an optimization problem under fairness constraints allows for their representation as bistochastic matrices which satisfy the fairness notions \eqref{eq:fairness_of_exposure} exactly. 
However, this method suffers {\bf limitation C} as it 
requires to solve a linear programming problem 
at each iteration of training and at inference, whose number of variables in $\mathbf{\Pi} \in \mathbb{R}^{n \times n}$ scales quadratically as $\cO(n^2)$ becoming prohibitively large as the item list grows. Additionally, at inference time, the policy must be sampled to produce rankings; this requires a Birhoff-Von Neumann (BVM) decomposition of the matrix $\mathbf{\Pi}$ into a convex combination of permutation matrices, which is also expensive when $n$ is large \citep{singh2018fairness}. Finally, in the case of multiple groups ($m>2$), the fairness constraints 
can become infeasible, making this formulation unwieldly (\textbf{limitation B}). An extended review of related work is provided in Section \ref{sec:related_work}.

Figure \ref{fig:variance} shows the query-level fairness violations due to each method discussed in this section, where fairness parameters in each case are increased maximally without substantially compromising utility. In addition to higher average violations, penalty-based methods \citep{zehlike2020reducing,singh2019policy} also lead to prevalence of outliers.
These three existing fair LTR methods are used as baselines for comparison in Section \ref{sec:experiments}. The SOFaiR framework proposed next most resembles \citep{kotary2022end}, as it combines learning of relevance scores end-to-end with constrained optimization. At the same time, it aims to improve over \citep{kotary2022end} by addressing the three main limitations stated above. By integrating an alternative optimization component with its predictive model, SOFaiR can achieve faster runtime, and avoid the BVM decomposition at inference time, while naturally accommodating fairness over an arbitrary number of groups. 

\section{Smart OWA Optimization for Fair Learning to Rank (SOFaiR)}
\label{sec:overview}

This section provides an overview of the proposed SOFaiR framework for learning fair ranking policies that overcomes limitations A, B, and C. Sections \ref{sec:forward_pass} and \ref{sec:backward_pass} will then detail the core solution approaches required to incorporate
its proposed fair ranking optimization module into efficient, end-to-end trainable fair LTR models.
As illustrated in Figure \ref{fig:pipeline}, SOFaiR's core concept is to 
integrate the learning of relevance scores with a module which optimizes fair ranking policies in-the-loop. By doing so it achieves a favorable balance of fairness and utility relative to other in-processing methods. The key difference in its approach relative to \citep{kotary2022end} is in the design of its optimization model which leverages Ordered Weighted Average (OWA) objectives (reviewed next) to enforce fairness of exposure. 
By avoiding the imposition of fairness of exposure (see Equation \eqref{eq:fairness_of_exposure}) as a set of hard constraints on the optimization as in \citep{kotary2022end,singh2018fairness}, it maintains the simple feasible region $\mathbf{\Pi} \in \cB$, over which efficient Frank-Wolfe based solution methods can be employed to optimize its OWA objective function as described in Section \ref{sec:forward_pass}. In turn, the particular form of the OWA optimization model in-the-loop necessitates a novel technique for its backpropagation, detailed in Section \ref{sec:backward_pass}. The OWA aggregation and its fairness properties in optimization problems are introduced next, followed by its role in the SOFaiR learning framework.

\subsection{Ordered Weighted Averaging Operator}
\label{sec:OWAoperator}

The $\textit{Ordered Weighted Average}$ (OWA) operator \citep{YAGER199380} has found applications in various decision-making fields \citep{10.5555/2464909} as a means of fairly aggregating multiple objective criteria. Let $\mathbf{x} \in \mathbb{R}^m$ be a vector of $m$ distinct criteria, and $\tau: \mathbb{R}^m \rightarrow \mathbb{R}^m$ be the sorting map for which $\tau(\mathbf{x}) \in \mathbb{R}^m$ holds the elements of $\mathbf{x}$ in increasing order. Then for any $\mathbf{w}$ satisfying $\{ \mathbf{w} \in \mathbb{R}^m: \sum_i w_i = 1, \mathbf{w} \geq 0 \}$, the OWA aggregation with weight $\mathbf{w}$ is defined as a linear functional on  $\tau(\mathbf{x})$:
\begin{equation}
\label{eq:OWA_definition}
 \textsc{OWA}_{\mathbf{w}}(\mathbf{x}) = \mathbf{w}^T \tau(\mathbf{x}),
 \end{equation}
which is convex and piecewise-linear in $\mathbf{x}$ \citep{ogryczak2003solving}. The so-called Generalized Gini Functions, or Fair OWA, are those for which the OWA weights $w_1 > w_2 \ldots > w_n$ are decreasing. Fair OWA functions possess the following three key properties for fairness in optimizing multiple criteria \citep{ogryczak2003solving}. 
{\bf (1)} \emph{Impartiality} means that all criteria are treated equally, in the sense that $\textsc{OWA}_{\mathbf{w}}(\mathbf{x}) = \textsc{OWA}_{\mathbf{w}}(\mathbf{x}_{\sigma})$ for any $\sigma \in \cP_{m}$. 
{\bf (2)} \emph{Equitability} is the property that marginal transfers from a criterion with higher value to one with lower value results in an increase in aggregated OWA value. That is , when $x_i > x_j +\epsilon$ and letting $\mathbf{x}_{\epsilon} = \mathbf{x}$ except at positions $i$ and $j$ where $(\mathbf{x}_{\epsilon})_i = \mathbf{x}_i - \epsilon$ and $(\mathbf{x}_{\epsilon})_j = \mathbf{x}_j + \epsilon$, it holds that $\textsc{OWA}_{\mathbf{w}}(\mathbf{x}_\epsilon) >  \textsc{OWA}_{\mathbf{w}}(\mathbf{x})$.  
{\bf (3)} \emph{Monotonicity} means that $\textsc{OWA}_{\mathbf{w}}(\mathbf{x})$ is an increasing function of each element of $\mathbf{x}$. The monotonicity property implies that solutions which optimize \eqref{eq:OWA_definition} are Pareto Efficient solutions of the underlying multiobjective problem, thus that no single criteria can be raised without reducing another \citep{ogryczak2003solving}. 
Taken together, it is known that maximization of aggregation functions which satisfy these three properties produces so-called \emph{equitably efficient solutions}, which possess the main intuitive properties needed for a solution to be deemed "fair"; see \citep{kostreva1999linear} for a formal definition. As shown next, the SOFaiR framework ensures group fairness by leveraging a fair OWA aggregation of group exposures $\textsc{OWA}_{\mathbf{w}}(\cE_G (\mathbf{\Pi}))$ in the objective function of its integrated fair ranking optimization module.

\subsection{End-to-End Learning in SOFaiR}
As illustrated in Figure \ref{fig:pipeline}, the SOFaiR framework uses a prediction model $\cM_{\mathbf{\theta}}$ with learnable weights $\mathbf{\theta}$, which produces relevance scores $\hat{\bm{y}}_q$ from a list of item features $\mathbf{x}_q$. Its key component is an optimization module which maps the prediction $\hat{\bm{y}}_q$ to an associated ranking policy $\bm{\Pi}^{\star}(\hat{\bm{y}}_q)$. The following optimization problem defines $\bm{\Pi}^{\star}(\hat{\bm{y}}_q)$ as the ranking policy which optimizes a trade-off between fair OWA aggregation of group exposures with the expected DCG (as per Equation \eqref{eq:birkhoff_dcg_work}) under relevance scores $\hat{\bm{y}}_q$. In SOFaiR, it defines, for any chosen weight $0 \leq \lambda \leq 1$, a mapping which can be viewed akin to a neural network layer, representing the last layer of $\cM$:
\begin{equation}
    \label{model:OWA_fair_rank}
    \bm{\Pi}^{\star}(\hat{\bm{y}}_q) = {\argmax}_{\bm{\Pi}  \in \cB} \;\; 
     (1 - \lambda) \cdot u(\bm{\Pi},\hat{\bm{y}}_q) +  \lambda \cdot  \textsc{OWA}_{\bm{w}}(\cE_G (\bm{\Pi})),
\end{equation}
wherein the Birkhoff Polytope $\cB$ is the set of all bistochastic matrices, as defined in  Equation \eqref{model:birkhoff_polytope}. 
Let the objective function of \eqref{model:OWA_fair_rank} be named $f(\mathbf{\Pi},\hat{\bm{y}}_q )$. It is a convex combination of two terms measuring user utility and fairness, whose trade-off is controlled by a single coefficient $0 \leq \lambda \leq 1$. The former term measures expected user utility $u(\bm{\Pi},\hat{\bm{y}}_q) = \hat{\bm{y}}_q^\top \, \bm{\Pi} \, \bm{b}$, while the latter term measures OWA aggregation of the group exposures. It is intuitive to see that when $\lambda = 1$, the optimization \eqref{model:OWA_fair_rank} returns a ranking policy that minimizes disparities in group exposure, without regard for relevance. When $\lambda = 0$, it returns a deterministic policy which ranks the items in order of the estimated scores $\hat{\bm{y}}_q$. Intermediate values $0 < \lambda < 1$ result in policies which trade off the effects of each term, balancing utility and fairness to various degrees. As $\lambda$ increases, disparity between the exposure of protected groups must decrease; this leads to a practical mechanism for achieving a desired level of fairness with minimal compromise to utility.

Since Equation \eqref{model:OWA_fair_rank} defines a direct mapping from $\hat{\bm{y}}_q$ to $\bm{\Pi}^{\star}(\hat{\bm{y}}_q)$, the problem of learning fair ranking policies reduces to a problem of learning relevance scores. This corresponds to estimating the objective function $f$ via its missing coefficients $\bm{y}_q$. The SOFaiR training method defines a loss function between predicted and ground-truth relevance scores, as the loss of optimality in $\mathbf{\Pi}^{\star}(\hat{\bm{y}}_q)$ with respect to objective $f$ under ground-truth $\bm{y}_q$, caused by prediction error in $\hat{\bm{y}}_q$. That is, the training objective is to minimize \emph{regret} in $f$ induced by $\hat{\bm{y}}_q$, defined as:
\begin{equation}
    \label{eq:regret}
    \textit{regret}(\hat{\bm{y}}_q,\bm{y}_q) = f(\;\Pi^{\star}(\bm{y}_q), \bm{y}_q\;) - f(\;\Pi^{\star}(\hat{\bm{y}}_q), \bm{y}_q\;)  . 
\end{equation}
The composition $\mathbf{\Pi}^{\star} \circ \cM_{\theta}$ defines an integrated prediction and optimization model which maps item features to fair ranking policies. Training the integrated model by stochastic gradient descent follows these steps in a single iteration:
\begin{enumerate} 
    \item   For sample query $q$ and item features $\textbf{x}_q$, a predictive model $\cM_{\mathbf{\theta}}$ produces estimated relevance scores $\hat{\mathbf{y}}_q$ .
    \item The predicted scores $\hat{\mathbf{y}}_q$ are used to populate the unknown parameters of an optimization problem \eqref{model:OWA_fair_rank}. A solution  algorithm is employed to find $\Pi^*(\hat{\bf{y}}_q)$, the optimal fair ranking policy relative to $\hat{\mathbf{y}}_q$.
    \item The regret loss \eqref{eq:regret} is backpropagated through the calculations of steps $(1)$ and $(2)$, in order to update the model weights $\mathbf{\theta}$ by a gradient descent step.
\end{enumerate}
The following sections detail the main solution schemes for implementing steps $(2)$ and $(3)$. 
Section \ref{sec:forward_pass} shows how recently proposed fair ranking optimization techniques from \citep{do2022optimizing} can be adapted to the setting of this paper, in which fair ranking policies must be learned from empirical data. From this choice of optimization design arises a novel challenge in the backpropagation step $(3)$, since no known work has shown how to backpropagate the regret of a highly discontinuous OWA optimization program. 
Section \ref{sec:backward_pass} shows how to efficiently backpropagate the regret due to problem \eqref{model:OWA_fair_rank} for end-to-end learning. Then, Section \ref{sec:experiments} evaluates the SOFaiR framework against several other methods for learning fair ranking policies, on a set of benchmark tasks from the web search domain. 
\section{Forward Pass Optimization}
\label{sec:forward_pass}
\begin{figure}
    \centering
\includegraphics[width=0.9\textwidth]{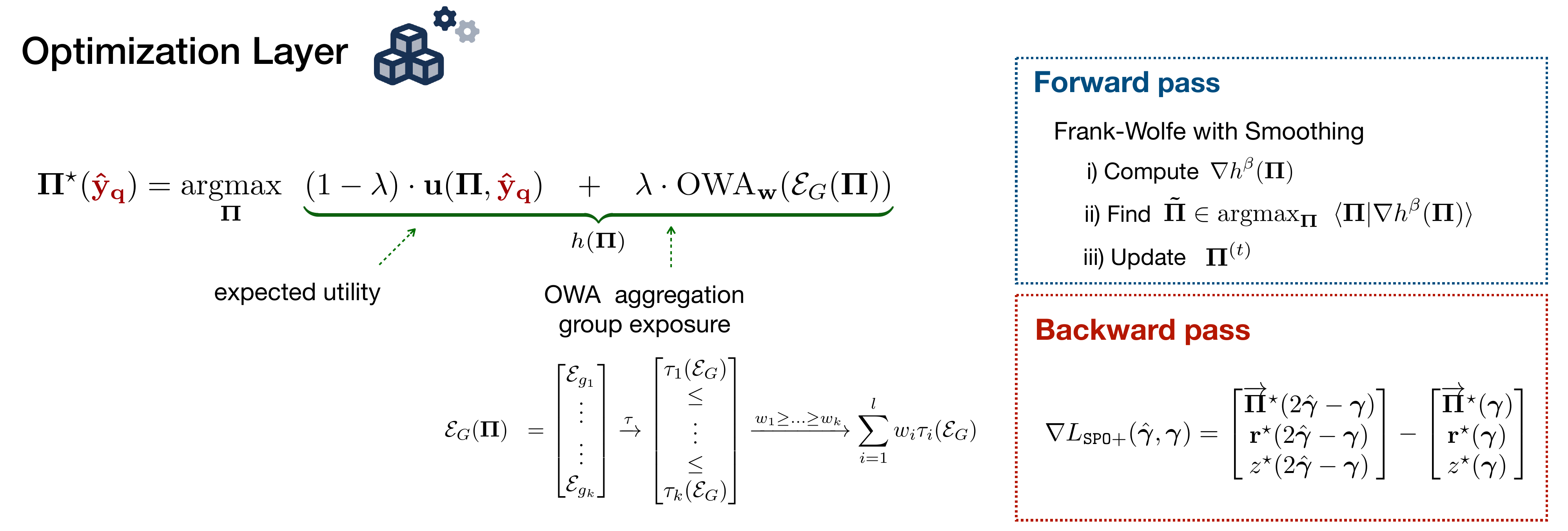}
    \caption{The differentiable optimization module employed in SOFaiR. It forward pass solves the problem \eqref{model:OWA_fair_rank} by an efficent Frank-Wolfe method. Its backward pass calculates the SPO+ subgradient, relative to its equivalent, but intractably large LP form.  }
    \label{fig:optim_layer}
\end{figure}
The main motivation for the formulation \eqref{model:OWA_fair_rank} of SOFaiR's fair ranking optimization layer is to render the optimization problem efficiently solvable. Its main exploitable attribute is its feasible region $\mathbf{\Pi}$, over which a \emph{linear} objective function can be quickly optimized by simply sorting a vector in $\mathbb{R}^n$, which has time complexity $n \log n$ \citep{cormen2022introduction}. This suggests an efficient solution by Frank-Wolfe methods, which solve a constrained optimization problem by a sequence of subproblems optimizing a linear approximation of the true objective function \citep{beck2017first}. This efficient solution pattern is made possible by the absence of additional group fairness constraints on the policy variable $\mathbf{\Pi}$.

Frank-Wolfe methods solve a convex constrained optimization problem $\argmax_{\mathbf{x} \in \mathbb{S}} f(\mathbf{x}) $ by computing the iterations 
\begin{equation}
    \label{eq:frank_wolfe_generic}
        \mathbf{x}^{(k+1)} = (1 - \alpha^{(k)})\mathbf{x}^{(k)} + \alpha^{(k)} \argmax_{\mathbf{y} \in \mathbb{S}} \langle \mathbf{y}, \nabla  f(\mathbf{x}^{(k)}) \rangle.
\end{equation}
Convergence to an optimal solution is guaranteed when $f$ is \emph{differentiable} and with $\alpha^{(k)} = \frac{2}{k+2}$ \citep{beck2017first}. However, the main obstruction to solving \eqref{model:OWA_fair_rank} by the method \eqref{eq:frank_wolfe_generic} is that $f$ in our case includes a \emph{non-differentiable} OWA function. A path forward is shown in \citep{lan2013complexity}, which shows convergence can be guaranteed by optimizing a smooth surrogate function $f^{(k)}$ in place of the nondifferentiable $f$ at each step of \eqref{eq:frank_wolfe_generic}, in such a way that the $f^{(k)}$ converge to the true $f$ as $k \to \infty$. 

It is proposed in \citep{do2022optimizing} to solve a two-sided fair ranking optimization with OWA objective terms, by the method of \citep{lan2013complexity}, where $f^{(k)}$ is chosen to be a Moreau envelope $h^{\beta_k}$ of $f$, a $\frac{1}{\beta_k}$-smooth approximation of $f$ defined as \citep{beck2017first}:
\begin{equation}
    \label{eq:moreau_env}
    h^{\beta}(\mathbf{x}) = \min_{\mathbf{y}} f(\mathbf{y}) + \frac{1}{2 \beta} \|   \mathbf{y} - \mathbf{x}  \|^2.
\end{equation}
When $f = \textsc{OWA}_{\bm{w}}$, let its Moreau envelope be denoted $\nabla \textsc{OWA}_{\bm{w}}^{\beta}$;
it is shown in \citep{do2022optimizing} that its gradient can be computed as a projection onto the permutahedron induced by modified OWA weights $\tilde{\bm{w}} = -(w_m, \ldots, w_1)$. By definition, the permutahedron $\cC(\tilde{\bm{w}}) = \textsc{conv}(\{ \bm{w}_{\sigma}: \forall \sigma \in \cP_m \})$ induced by a vector $\tilde{\bm{w}}$ is the convex hull of all its permutations. In turn, it is shown in \citep{blondel2020fast} that the permutahedral projection $\nabla \textsc{OWA}_{\bm{w}}^{\beta}(\mathbf{x}) = \proj_{\cC(\tilde{\bm{w}})} (\nicefrac{\bm{x}}{\beta})$ can be computed in $m \log m$ time as the solution to an isotonic regression problem using the Pool Adjacent Violators algorithm. To find the overall gradient of $\textsc{OWA}_{\bm{w}}^{\beta}$ with respect to optimization variables $\mathbf{\Pi}$, a convenient form can be derived from the chain rule:
\begin{equation}
    \nabla_{\bm{\Pi}} \; \textsc{OWA}_{\bm{w}}^{\beta}(\cE(\bm{\Pi}))= \bm{\mu} \bm{b}^T.
\end{equation}
where $\bm{\mu} = \proj_{\cC(\tilde{\bm{w}})} (\nicefrac{\cE(\mathbf{\Pi})}{\beta})$ and $\cE(\bm{\Pi})$ is the vector of all item exposures \citep{do2022optimizing}. For the case where group exposures $\cE_G(\bm{\Pi})$ are aggregated by OWA, first note that by Equation \ref{eq:group_exposure_closed}, $\cE_G(\bm{\Pi}) = \bm{A} \bm{\Pi} \bm{b}$, where $\bm{A}$   is the matrix composed of stacking together all group indicator vectors $\bm{1}_g \; \forall g \in G$. Since $\cE(\bm{\Pi}) = \bm{\Pi} \bm{b}$, this implies   $\cE_G(\bm{\Pi}) = \cE(\bm{A} \bm{\Pi})$, thus 
\begin{equation}
    \label{eq:OWA_moreau_grad}
    \nabla_{\bm{\Pi}} \; \textsc{OWA}_{\bm{w}}^{\beta}(\cE_G(\bm{\Pi}))= (\bm{A}^T \tilde{\bm{\mu}}) \; \bm{b}^T.
\end{equation}
by the chain rule, and where $\tilde{\bm{\mu}} = \proj_{\cC(\tilde{\bm{w}})} (\nicefrac{\cE_G(\bm{A} \mathbf{\Pi})}{\beta})$. It remains now to compute the gradient of the user relevance term $u(\bm{\Pi}, \hat{\bm{y}}_q) =  \hat{\bm{y}}_q^T \bm{\Pi} \; \bm{b}$ in Problem \ref{model:OWA_fair_rank}. As a linear function of the matrix variable $\bm{\Pi}$, its gradient is $\nabla_{\bm{\Pi}} \; u(\bm{\Pi}, \hat{\bm{y}}_q) =  \hat{\bm{y}}_q \; \bm{b}^T$, which is evident by comparing to the equivalent vectorized form $ \hat{\bm{y}}_q^T \bm{\Pi} \; \bm{b} = \overrightarrow{ \hat{\bm{y}}_q \; \bm{b}^T} \cdot \overrightarrow{\bm{\Pi}}$. Combining this with \eqref{eq:OWA_moreau_grad}, the total gradient of the objective function in \eqref{model:OWA_fair_rank} with smoothed OWA term is  $(1 - \lambda) \cdot \hat{\bm{y}}_q \; \bm{b}^T  +  \lambda \cdot     (\bm{A}^T\tilde{\bm{\mu}}) \; \bm{b}^T $, which is equal to $\left((1 - \lambda) \cdot  \hat{\bm{y}}_q +  \lambda \cdot (\bm{A}^T\tilde{\bm{\mu}}) \right)\; \bm{b}^T$. Therefore the SOFaiR module's Frank-Wolfe linearized subproblem is
\begin{equation}
    \label{eq:sofair_subproblem}
    \argmax_{\bm{\Pi} \in \cB}  \left\langle \bm{\Pi}, \left((1 - \lambda) \cdot  \hat{\bm{y}}_q +  \lambda \cdot (\bm{A}^T\tilde{\bm{\mu}}) \right)\; \bm{b}^T   \right\rangle
\end{equation}

\begin{algorithm}[!t]
\caption{Frank-Wolfe with Moreau Envelope Smoothing to solve \eqref{model:OWA_fair_rank}}\label{alg:FWS}
\KwInput{predicted relevance scores $\bm{\hat{y}} \in \RR^n$, group mask $\bm{A}$, max iteration T, smooth seq. $(\beta_k)$}
\KwOutput{ranking policy $\bm{\Pi}^{(T)} \in \RR^{ n \times n}$}

Initialize $\bm{\Pi}^{(0)}$ as $\bm{P} \in \cP$ which sorts $\bm{{\hat{y}}}$ in decreasing order\;
\For{$k = 1, \ldots, T$}
{
    $\tilde{\bm{\mu}} \gets \proj_{\cC(\tilde{\bm{w}})} (\nicefrac{\cE_G(\bm{A} \mathbf{\Pi})}{\beta_k})$\;
    
    $\hat{\bm{\mu}} \gets (1 - \lambda) \cdot  \hat{\bm{y}}_q +  \lambda \cdot (\bm{A}^T\tilde{\bm{\mu}}) $\;
    
    $\hat{\sigma} \gets \textit{argsort} (-\hat{\bm{\mu}})$\;
    
    Let $\bm{P}^{(k)} \in \mathcal{P}$ such that $\bm{P}^{(k)}$ represents $\hat{\sigma}$\;
    
    $\bm{\Pi}^{(k)} \gets \frac{k}{k+2} \bm{\Pi}^{(k-1)} + \frac{2}{k+2} \bm{P}^{(k)} $\;
}
Return $\bm{\Pi}^{(T)}$\;
\end{algorithm}

To implement the Frank-Wolfe iteration \eqref{eq:frank_wolfe_generic}, this linearized subproblem should have an efficient solution. To this end, the form of each gradient above as a cross-product of some vector with the position biases $\bm{b}$ can be exploited. Note that as the expected DCG under relevance scores $\bm{y}$, the function $\bm{y}^T \Pi \; \bm{b}$ is maximized by the permutation matrix $\bm{P} \in \cP_n$ which sorts the relevance scores $\bm{y}$ decreasingly. But since $ \bm{y}^T \bm{\Pi} \; \bm{b} = \overrightarrow{ \bm{y} \; \bm{b}^T} \cdot \overrightarrow{\bm{\Pi}}$, we identify $ \bm{y}^T \bm{\Pi} \; \bm{b}$ as the linear function of $\overrightarrow{\bm{\Pi}}$ with gradient $\overrightarrow{ \bm{y} \; \bm{b}^T}$. Therefore problem \eqref{eq:sofair_subproblem} can be solved in $\cO(n \log n)$, simply by finding $\bm{P} \in \cP_n$ as the argsort of the vector $ ((1 - \lambda) \cdot  \hat{\bm{y}}_q +  \lambda \cdot (\bm{A}^T\tilde{\bm{\mu}}) )$ in decreasing order. A more formal proof, cited in \citep{NEURIPS2021_48259990}, makes use of \citep{hardy1952inequalities}.

The overall method is presented in Algorithm \ref{alg:FWS}. Decay of the smoothing parameter $\beta_t = \frac{\beta_0}{\sqrt{t}}$ satisfies the conditions for convergence stated in \citep{lan2013complexity} when $\beta_0$ is sufficiently large. Sparse matrix additions each require $\cO(n)$ operations, so that  Algorithm \ref{alg:FWS} maintains $\cO(n \log n)$ complexity per iteration. An important advantage of Algorithm \ref{alg:FWS} over the fair ranking LP employed in SPOFR \citep{kotary2022end}, is that the solution iterates $\bm{P}^{(k)}$ automatically provide a decomposition of the policy matrix $\bm{\Pi} = \rho_k \bm{P}^{(k)}$ as a convex combination of rankings, by which it can be readily sampled as a discrete probability distribution. In contrast, the LP module used in SPOFR \citep{kotary2022end} provides as its solution only a matrix $\bm{\Pi} \in \cB$, which must be decomposed using the Birkhoff Von Neumann decomposition, adding substantially to its total runtime.

\section{Backpropagation}
\label{sec:backward_pass}
The formulation of the optimization module \eqref{model:OWA_fair_rank} allows for efficient solution via Algorithm \ref{alg:FWS}, but gives rise to a novel challenge in backpropagating the regret loss function through $\Pi^{\star}(\hat{\bm{y}}_q)$. By including an OWA aggregation of group exposure, its objective function is nonlinear and nondifferentiable. This section shows how to train the integrated prediction and OWA optimization model $\mathbf{\Pi}^{\star} \circ \cM_{\theta}$ to minimize the regret loss \eqref{eq:regret}, despite this challenge. As a starting point, we recognize the existing literature on "Predict-Then-Optimize" frameworks \citep{kotary2021end,mandi2023decision} for minimizing the regret due to prediction error in the objective coefficients of a linear program, denoted $\bm{c}$ below:
\begin{equation}
    \label{model:opt_generic_linear}
    \mathbf{x}^{\star}(\mathbf{c}) = \argmin_{\mathbf{A} \mathbf{x} \leq \mathbf{b}} \mathbf{c}^T \mathbf{x}. 
\end{equation}
Several known methods have been proposed \citep{elmachtoub2020smart, vlastelica2020differentiation, wilder2019melding, berthet2020learning} and well-established in the literature \citep{mandi2023decision} for end-to-end training of combined prediction and optimization models employing \eqref{model:opt_generic_linear}. Due to its OWA objective term, the fair ranking module \eqref{model:OWA_fair_rank} does not satisfy the LP form \eqref{model:opt_generic_linear} for which the aforementioned methods are taylored. The implementation of SOFaiR described here uses the "Smart Predict-Then-Optimize" (SPO) approach \citep{elmachtoub2020smart}, since its simple backpropagation rule requires only a solution to \eqref{model:opt_generic_linear} using a blackbox solution oracle. This allows its adaptation to the OWA optimization setting by constructing (but not solving) an equivalent but intractable linear programming form to \eqref{model:OWA_fair_rank}, as shown next.

\smallskip\noindent\textbf{End-to-End learning with SPO$+$ Loss.}  Viewed as a loss function, the regret \eqref{eq:regret} in solutions to problem \eqref{model:opt_generic_linear} is nondifferentiable and discontinuous with respect to predicted coefficients $\hat{\bm{c}}$, since solutions $\bm{x}^{\star}(\bm{c})$ must occur at one of finitely many vertices in $\mathbf{A} \mathbf{x} \leq \mathbf{b}$. The SPO$+$ loss function proposed in \citep{elmachtoub2020smart} is by construction a Fischer-consistent, \emph{subdifferentiable}  upper bound on regret. In particular, it is shown in \citep{elmachtoub2020smart} that 
\begin{equation}
\label{model:SPO_loss}
 L_{\texttt{SPO}+}(\hat{\bm{c}},\bm{c}) = \max_{\mathbf{x}}(\mathbf{c}^T \mathbf{x} - 2\hat{\mathbf{c}}^T \mathbf{x}) + 2 \hat{\mathbf{c}}^T \mathbf{x}^{\star}(\mathbf{c}) - \mathbf{c}^T \mathbf{x}^{\star}(\mathbf{c}),
\end{equation}
possesses these properties, and a subgradient at $\hat{\mathbf{c}}$ is
\begin{equation}
\label{model:SPO_grad}
 \nabla L_{\texttt{SPO}+}(\hat{\bm{c}},\bm{c}) =\bm{x}^{\star}(2\hat{\bm{c}} - \bm{c})  - \bm{x}^{\star}(\bm{c})  .
\end{equation}
Minimizing the surrogate loss \eqref{model:SPO_loss} by gradient descent using \eqref{model:SPO_grad} is key to minimizing the solution regret in problem \eqref{model:opt_generic_linear} due to error in a predictive model which predicts the parameter $\mathbf{c}$.

\smallskip\noindent\textbf{SPO+ loss in SOFaiR.}
We now show how the SPO training framework described above for the problem type \eqref{model:opt_generic_linear} can be used to efficiently learn optimal fair policies in conjunction with problem \eqref{model:OWA_fair_rank}. The main idea is to derive an SPO+ subgradient for regret in \eqref{model:OWA_fair_rank}, through an equivalent linear program \eqref{model:opt_generic_linear}, but without solving it as such. This is made possible by the fact that the subgradient \eqref{model:SPO_grad} can be expressed as a difference of two optimal solutions, which can be furnished by any optimization oracle which solves the mapping \eqref{model:OWA_fair_rank}, which includes Algorithm \eqref{alg:FWS}. 

First note, as it is shown in \citep{ogryczak2003solving}, that the OWA function \eqref{eq:OWA_definition} can be expressed as
\begin{equation}
    \label{model:OWA_min_form}
         \textit{OWA}_{\mathbf{w}}(\mathbf{r}) = {\min}_{\sigma \in \mathcal{P}} \;\; \mathbf{w}_{\sigma} \cdot \mathbf{r},
\end{equation}
and as an equivalent linear programming problem which views the minimum inner product above as the maximum lower bound among all possible inner products with the permuted OWA weights: 
\begin{subequations}
    \begin{align}
    \label{model:OWA_LP_form}
         \textit{OWA}_{\mathbf{w}}(\mathbf{r}) = {\max}_{z} \;\;\;\; & z \\
         \textit{s.t.}\;\;\;\; & z \leq \mathbf{w}_{\sigma} \cdot \mathbf{r}, \;\;\;\;\;\; \forall \sigma \in \cP,
    \end{align}
\end{subequations}
where $\cP$ contains all possible permutations of $\left[ n \right]$ when $\mathbf{w} \in \mathbb{R}^n$. This allows SOFaiR's OWA optimization model \eqref{model:OWA_fair_rank} to be recast in a linear programming form using auxiliary optimization variables $\mathbf{r}$ and $z$:
\begin{subequations}
    \label{model:OWA_fair_rank_LP_form}
    \begin{align}
    \label{eq:OWA_fair_rank_LP_form_obj}
    (\bm{\Pi}^{\star}, \bm{r}^{\star}, z^{\star})(\hat{\bm{y}}_q) = {\argmax}_{\bm{\Pi} \in \cB,\;\mathbf{r},\;z} &\;\; 
     (1 - \lambda) \cdot \hat{\bm{y}}_q^\top \, \bm{\Pi} \, \bm{b} +  \lambda \cdot  z  \\
    \mbox{subject to:} & \;\;\; z \leq \bm{w}_{\sigma} \cdot r,  \;\;\;\;\;\; \forall \sigma \in \cP \label{eq:OWA_fair_rank_LP_LB}\\
    & \;\;\;  r = \cE_G (\mathbf{\Pi}).
    \end{align}
\end{subequations}

According to \citep{ogryczak2003solving}, this alternative LP form of OWA optimization is mostly of theoretical significance, since the set of constraints \eqref{eq:OWA_fair_rank_LP_LB} grows factorially in the size of $\mathbf{r}$, one for each possible permutation thereof. This makes     \eqref{model:OWA_fair_rank_LP_form} impractical for computing a solution to the original OWA problem \eqref{model:OWA_fair_rank}, which we instead solve by Algorithm \ref{alg:FWS}. On the other hand, we show that problem \eqref{model:OWA_fair_rank_LP_form} \emph{is} practical for deriving a \emph{backpropagation} rule through the OWA problem \eqref{model:OWA_fair_rank}.

Since the unknown parameters $\hat{\bm{y}}_q$ appear only in its linear objective function, this parametric LP problem \eqref{model:OWA_fair_rank_LP_form} fits the form \eqref{model:opt_generic_linear} required for training with SPO+ subgradients. To derive the subgradient explicitly, rewrite the linear objective term $\hat{\bm{y}}_q^\top \, \bm{\Pi} \, \bm{b} \; = \; \overrightarrow{\hat{\bm{y}}_q \; \bm{b}^T } \; \cdot \;  \overrightarrow{\bm{\Pi}} $. Then in terms of the augmented variables $(\mathbf{\Pi}, \mathbf{r}, z)$, the objective function \eqref{eq:OWA_fair_rank_LP_form_obj} is 
\begin{equation}
    (1 - \lambda)_{x} \cdot \hat{\bm{y}}_q^\top \, \bm{\Pi} \, \bm{b} + \lambda \cdot z = {\underbrace{ \left[ \begin{matrix} (1-\lambda) \overrightarrow{\hat{\bm{y}}_q \; \bm{b}^T } \\ \mathbf{0} \\ 
    \lambda \end{matrix} \right]}_{\hat{\bm{\gamma}}}}^T  \left[ \begin{matrix} \overrightarrow{\bm{\Pi}}  \\ \mathbf{r} \\ z \end{matrix} \right].
\end{equation}
Now the SPO+ loss subgradient can be readily expressed with respect to the augmented scores $\hat{\bm{\gamma}}$ defined as above:

\begin{equation}
\label{model:SPO_grad_SOFaiR}
 \nabla L_{\texttt{SPO}+}(\hat{\bm{\gamma}},\bm{\gamma}) = \left[ \begin{matrix} \overrightarrow{\bm{\Pi}}^{\star}(2\hat{\bm{\gamma}} - \bm{\gamma})  \\ \mathbf{r}^{\star}(2\hat{\bm{\gamma}} - \bm{\gamma}) \\ z^{\star}(2\hat{\bm{\gamma}} - \bm{\gamma}) \end{matrix} \right]  - \left[ \begin{matrix} \overrightarrow{\bm{\Pi}}^{\star}(\bm{\gamma})  \\ \mathbf{r}^{\star}(\bm{\gamma}) \\ z^{\star}(\bm{\gamma}) \end{matrix} \right]  ,
\end{equation}
using \eqref{model:SPO_grad}, and where $\bm{\gamma}$ is the augmented score based on ground-truth $\mathbf{y}_q$. Finally, backpropagation from $\hat{\bm{\gamma}}$ to the base prediction $\hat{\mathbf{y}}_q = \cM_{\bm{\theta}}(\mathbf{x}_q)$ is performed by automatic differentiation, and likewise from $\hat{\mathbf{y}}_q$ to the model weights $\bm{\theta}$.

Both terms in \eqref{model:SPO_grad_SOFaiR} can be produced by using Algorithm \ref{alg:FWS} to solve \eqref{model:OWA_fair_rank} for $\mathbf{\Pi}^{\star}$. Then, the remaining variables $\mathbf{r}^{\star}$ and $z^{\star}$ are easily completed as groups exposures $\mathbf{r} = \cE_G(\Pi^{\star})$ and their associated OWA value $z$, respectively. Importantly, the rightmost term of \eqref{model:SPO_grad_SOFaiR} is independent of any prediction; therefore it is \emph{precomputed} in advance of training. Thus, backpropagation using \eqref{model:SPO_grad_SOFaiR} consists of computing the difference between two solutions, one of which comes from the forward pass and while the other is precomputed before training. The complexity of this backward pass consists of $\mathcal{O}(n^2)$ subtractions, which grows only linearly in the size of the matrix variable $\mathbf{\Pi} \in \mathbb{R}^{n \times n}$. The differentiable fair ranking optimization module of SOFaiR, with its forward and backward passes, is summarized in Figure \ref{fig:optim_layer}.

\section{Experiments}
\label{sec:experiments}

Next we evaluate SOFaiR against two prior in-processing methods \citep{singh2019policy,zehlike2017fa}, and the end-to-end framework \citep{kotary2021end}, denoted as FULTR, DELTR, and SPOFR, respectively.
We assess the performance on two datasets:
\begin{itemize}
    \item Microsoft Learn to Rank (MSLR) is a standard benchmark for LTR with queries from Bing and manually-judged relevance labels. It includes 30,000 queries, each with an average of 125 assessed documents and 136 ranking feature. Binary protected groups is defined using the 50th percentile of QualityScore attribute. For multi-group cases, group labels are defined using evenly-spaced quantiles. 
    \item  Yahoo! Learning to Rank Challenge (Yahoo LETOR) contains 19,944 queries and 473,134 documents with 519 ranking features. Binary protected groups is defined using feature id 9 following \citep{jia2021calibrating} and the 50th percentile as the threshold.
\end{itemize}
For MSLR, we randomly sample 10,000 queries for training and 1,000 queries each for validation and testing. We create datasets with varying list sizes (20, 40, 60, 80, 100 documents) for MSLR and (20, 40 documents) for Yahoo LETOR.

\smallskip\noindent{\bf Models and hyperparameters.}
A neural network (NN) with three hidden layers is trained using Adam Optimizer with a learning rate of 0.1 and a batch size of 256. The size of each layer is halved, and the output is a scalar item score. Results of each hyperparameter setting is are taken on average over five random seeds.

Fairness parameters, considered as hyperparameters, are treated differently. LTR systems aim to offer a trade-off between utility and group fairness, since the cost of increased fairness results in decreased utility. In DELTR, FULTR, and SOFaiR, this trade-off is indirectly controlled through the fairness weight, denoted as $\lambda$ in \eqref{eq:fair_ERM} and \eqref{model:OWA_fair_rank}. Larger values of $\lambda$ indicate more preference towards fairness. In SPOFR, the allowed violation \eqref{eq:fairness_violation} of group fairness is specified directly. Ranking utility and fairness violation are assessed using average DCG ( Equation \eqref{eq:utilty_def}) and fairness violation (Equation \eqref{eq:fairness_violation}), respectively. The metrics are computed as averages over the entire test dataset.

\subsection{Running Time Analysis} 

\begin{figure}
    \centering
    \includegraphics[width=0.8\textwidth]{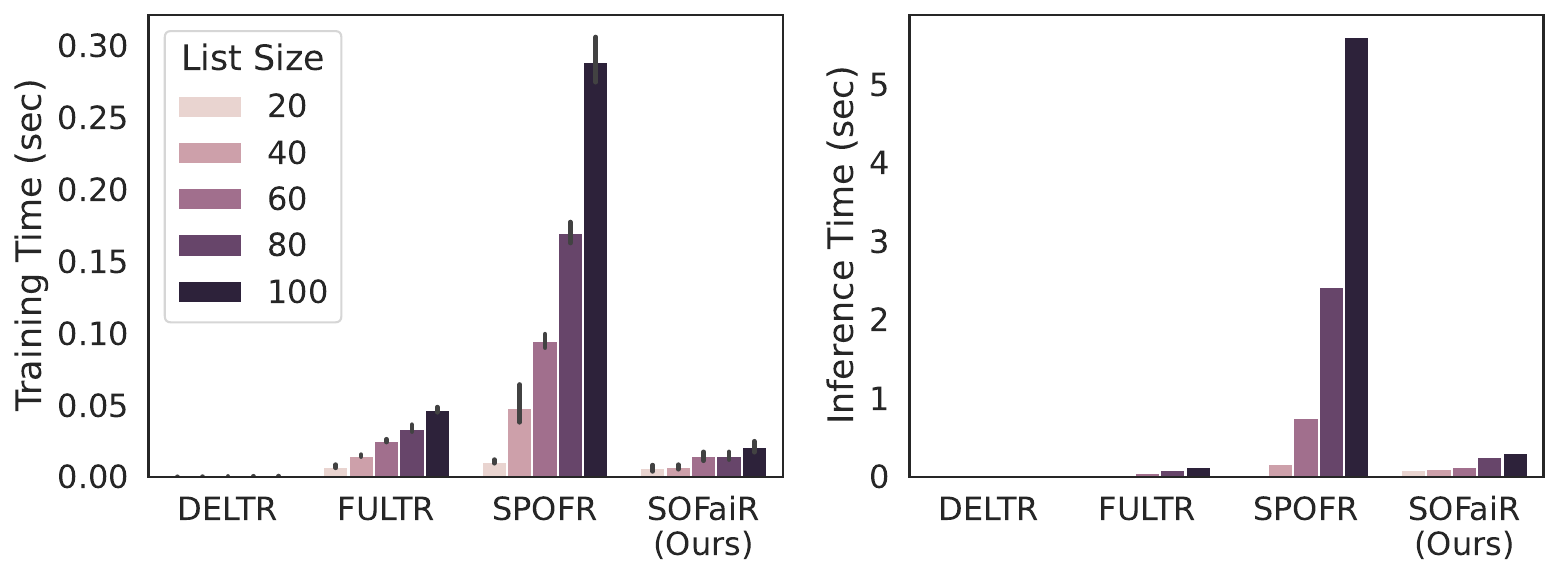}
    \vspace{-8pt}
    \caption{Running time benchmark on MSLR-Web10k dataset}
    \label{fig:benchmark}
\end{figure}

Our analysis begins with a runtime comparison between SOFaiR and other LTR frameworks, to show how it overcomes {\bf limitation C}, described in Section \ref{sec:limitations}. Figure \ref{fig:benchmark} shows the average training and inference time per query for each method, focusing on the binary group MSLR dataset across various list sizes. First notice the drastic runtime reduction of SOFaiR compared to SPOFR, both during training and inference. While SPOFR's training time exponentially increases with the ranking list size, SOFaiR's runtime increases only moderately, reaching over one order of magnitude speedup over SPOFR for large list sizes. Notably, the number of iterations of Algorithm \ref{alg:FWS} required for sufficient accuracy in training to compute SPO+ subgradients are found to less than those required for solution of \eqref{model:OWA_fair_rank} at inference. Thus the reported results use $100$ iterations in training and $500$ at inference.  Importantly, reported runtimes under-estimate the efficiency gained by SOFaiR, since its PyTorch \citep{paszke2017automatic} implementation in Python is compared against the highly optimized code implementation of Google OR-Tools solver \citep{perron2011operations}. DELTR and FULTR, as penalty-based methods, are more competitive in runtime. However, this comes at a cost of the achieved fairness level ({\bf limitation A}), as shown in the next section.

\subsection{Fairness and Utility Tradeoffs Analysis}

\begin{figure}
    \centering

    \includegraphics[width=0.475\textwidth]{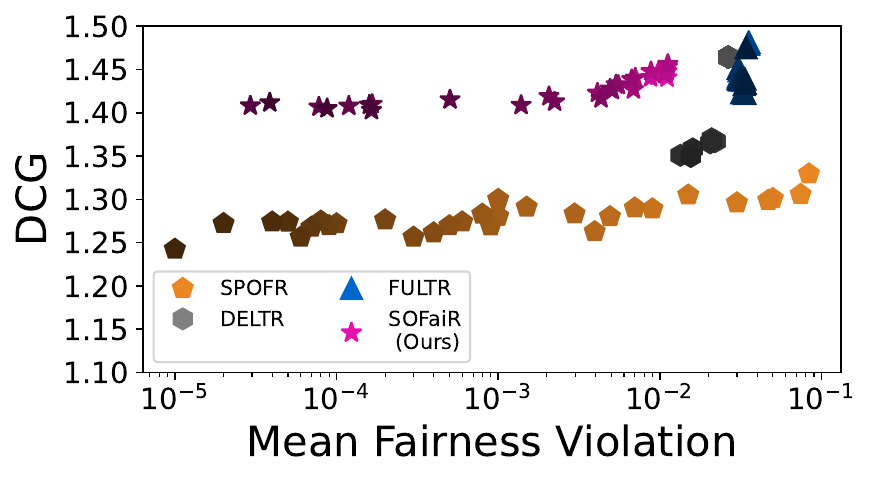}
    \includegraphics[width=0.475\textwidth]{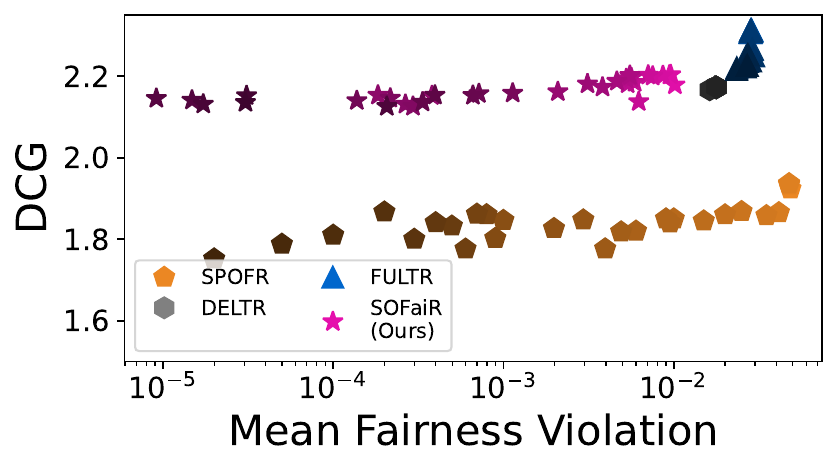}
    \includegraphics[width=0.475\textwidth]{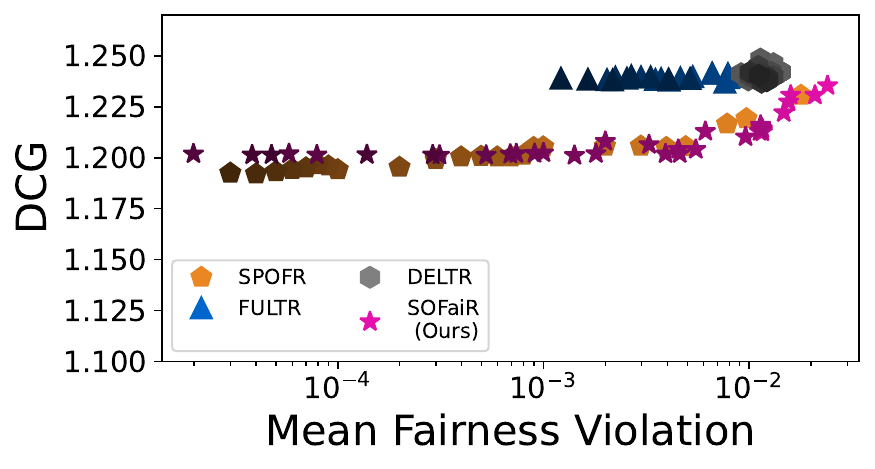}
    \includegraphics[width=0.475\textwidth]{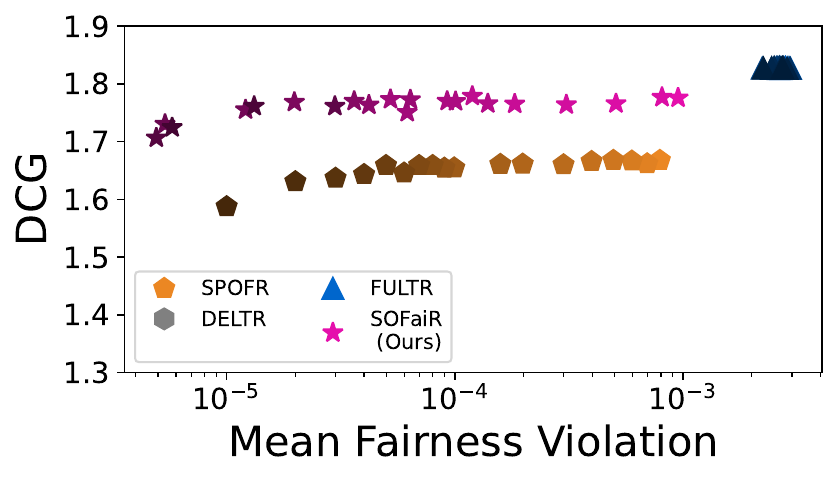}

    \caption{Benchmarking performance in term of fairnesss-utility trade-off on Yahoo-20 (top left), and Yahoo-40(top right). MSLR-20(bottom-left), MSLR-100 (bottom-right)}
    \label{fig:binary_trade_off}
\end{figure}

Next, we focus on comparing the utility and fairness of the various LTR frameworks analyzed. This section focuses on the two-group case, as none of the methods compared against was able to cope with multi-group case in our experiments (see next section). 
Figure \ref{fig:binary_trade_off} presents the trade-off between utility and fairness across the test sets for both Yahoo LETOR and MSLR datasets, encompassing their lowest and highest list sizes. For each method, the intensity of colors represents the magnitude of its fairness  parameter. A progression from lighter to darker colors indicates an increase in the importance placed on fairness. Consequently, darker colors are expected to correspond with more restrictive models, characterized by lower DCG scores (y-axis) but also fewer fairness violations (x-axis). Each point in the figure represents the largest DCG score obtained from a fairness hyperparameter search, as detailed in Appendix \ref{app:hyper}. Note that points on the grid that are higher on the y-axis and lower on the x-axis represent superior results. 



Firstly, notice that most points associated with methods DELTR and FULTR are clustered in a small region with both high DCG and (log-scaled) fairness violations. While these methods reach an order of magnitude reduction in fairness violation on some datasets, the effect is inconsistent, especially as the item list size increases ({\bf limitation A}). In contrast, the end-to-end methods (SPOFR and the proposed SOFaiR) reach much lower fairness violations, underlining their effectiveness of their optimization modules in enforcing the fairness constraint. 

Both DELTR and FULTR reach competitive utilities, but they consistently display relatively high fairness violations, underscoring their limitations in providing a fair ranking solution. SOFaiR shows competitive fairness and utility performance compared to SPOFR, with a marked advantage in utility on some datasets. SPOFR ensures fairness but at the expense of efficiency, whereas SOFaiR reaches similar fairness levels at a fraction of the required runtime. Additional results on datasets of various list sizes are included in Appendix \ref{app:benchmark}.

\subsection{Multi-Group Fairness Analysis}

\begin{figure}
    \centering
     \includegraphics[width=0.475\textwidth]{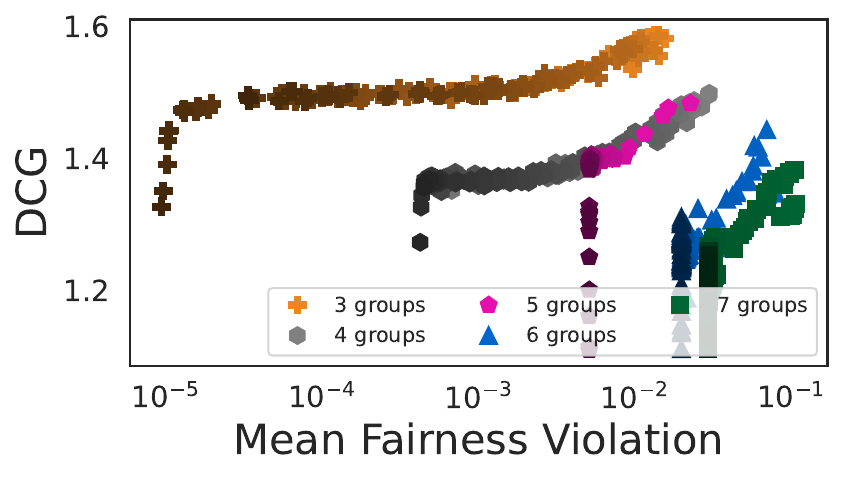}
  \includegraphics[width=0.475\textwidth]{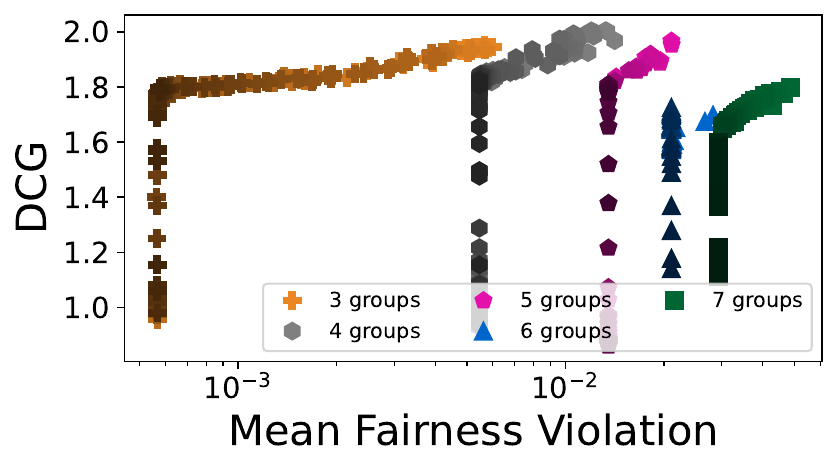}
    \caption{Fairness-utility tradeoff due to SOFaiR with multiple groups on MSLR-40 (left) and MSLR-100 (right) list size}
    \label{fig:multi_group}
\end{figure}

Finally, this section analyzes the fairness-utility trade-off in multi-group scenarios  using the SOFaiR framework. The SPOFR method returns infeasible solutions for most chosen fairness levels when multiple groups are introduced, preventing its evaluation on these datasets; this is naturally avoided in SOFaiR as the optimization of OWA aggregation  without constraints simply increases fairness to the extent feasible. While FULTR provides no code to evaluate multigroup fairness, its penalty function is in principle ill-equipped to handle multiple groups as it must scalarize all group fairness violations into a single loss function as mentioned in Section \ref{sec:limitations}  ({\bf limitation C}). 
Figure \ref{fig:multi_group} compares the average test DCG against the average fairness violation across various numbers of groups (ranging from 3 to 7) in the MSLR dataset, for list sizes of 40 and 100. Additional results for other list sizes in the MSLR dataset are available in \ref{app:multi}.

Each data point represents a single model's performance, with fairness parameters $\lambda$ adjusted between 0 and 1. Models prioritizing fairness show reduced fairness violations and lower utilities, indicated by darker colored points, compared to those with a lower emphasis on fairness, represented by lighter colored points.  A distinct trend is observed: as fairness parameters are relaxed, utility increases for all metrics and datasets. It is also evident that multi-group fairness comes at a higher cost to utility.  Predictably, saturation occurs in each curve, indicating that beyond a certain point, increasing the fairness weight does not further decrease fairness violations but merely reduces utility.

\section{Related Work}
\label{sec:related_work}
This paper is concerned with learning ranking models under fairness constraints, a problem for which a variety of methods have been developed in the web search domain. Unlike most previous works in this domain, its main solution framework follows the Predict-Then-Optimize (PtO) paradigm, in which optimization models are trained end-to-end with neural networks. This section surveys relevant works in the application area, before giving a brief overview of related works with respect to the PtO methodology.

\paragraph{\bf Learning Fair Ranking Policies.}
\label{sec:related_work_ltr}

Recent advancements in learning fair ranking policies fall into three main categories based on the incorporation of fairness criteria into the training pipeline. In the \emph{pre-processing} category, methods focus on mitigating bias in training data by transforming input data to make representations similar \cite{26, 27}. For instance, \cite{26} aims for fairness by ensuring indistinguishable item pairs on non-sensitive attributes become nearly indistinguishable in their fair representation. On the other hand, \emph{post-processing} methods refine rankings or scores after training to meet fairness criteria \cite{survey}. Examples include \cite{59}, which utilizes optimal transport to post-process the model's output scores for a fair score distribution, and works like \cite{zehlike2017fa, zehlike2022, 19}, which enforce a minimum portion of protected members in the top k ranking using statistical testing and rule-based selection. While \cite{singh2018fairness} addresses different fairness criteria, its scalability is limited to a small item list due to optimization system complexity. In contrast, methods like \cite{zehlike2022} work with specific fairness definitions and can scale to large item lists, as demonstrated on LinkedIn search. Finally, \cite{do2022optimizing} proposes an efficient optimization method for solving a two-sided fair ranking problem. Post-processing methods guarantee fairness but may sacrifice accuracy in user relevance. Since post-processing methods model the fair ranking policy separately from the learning of relevance scores, they generally guarantee fairness satisfaction at the cost of accuracy in terms of user relevance. 

\emph{In-processing} methods aim for improved accuracy-fairness trade-offs by integrating fairness criteria into the LTR training loop \cite{survey}. Typically, these methods enhance the training loss function with penalties for fairness violations, allowing the models to strike a balance between accuracy and fairness. However, it is common for the fairness criteria to be imperfectly satisfied. In this context, \cite{zehlike2020reducing} and \cite{singh2019policy} focus on equal group exposures, ensuring visibility of items at lower rankings. Additionally, \cite{4} employs pairwise accuracy to assess if candidates at higher ranks have higher relevant scores, while \cite{bower21, pmlr-v81-kamishima18a} guarantee a similar ranking policy for queries or items that differ in sensitive attributes.

\paragraph{\bf End-to-End Prediction and Optimization.}
\label{sec:related_work_pto}
Recent literature has been developed around constrained optimization models that are trained end-to-end with machine learning models \cite{kotary2021end}. In the Predict-Then-Optimize setting, a machine learning model predicts the unknown coefficients of an optimization problem. Then, backpropagation through the optimal solution of the resulting problem allows for end-to-end training of its objective value, under ground-truth coefficients, as a loss function. The primary challenge is backpropagation through the optimization model, for which a variety of alternative techniques have been proposed. When the optimization program is differentiable, this can be done through direct differentiation \cite{agrawal2019differentiable,agrawal2019differentiating,amos2017optnet,kotary2023folded}. Otherwise, various smoothing, randomization and approximation techniques are employed \cite{vlastelica2020differentiation, berthet2020learning, elmachtoub2020smart}. The main advantage of this framework is in achieving higher downstream objective function values, when compared to learning the unknown coefficients directly by regression to the ground-truth values.

Most similar to the present paper, \cite{kotary2022end} proposes a fair learning to rank method based on Predict-Then-Optimize with the fair ranking optimization model of \cite{singh2018fairness}, whose unknown coefficients are the relevance scores, while the loss function is user relevance via discounted cumulative gain. The main advantage of \cite{kotary2022end} is that it inherits the fairness guarantees of \cite{singh2018fairness} at the level of each query, along with precise control over fair-utility tradeoffs by setting the allowed fairness violation. By bringing a post-processing method \cite{singh2018fairness} into the training loop, it also shows considerable gains in user relevance over the alternative in-processing methods described above. However, this comes at a considerable computational cost of solving a large optimization problem for each sample at each iteration of training.

\section{Conclusions}
This paper presented SOFaiR, a method that employs an Ordered Weighted Average (OWA) optimization model to integrate fairness considerations into ranking processes. Its integration of constrained optimization  in an end-to-end differentiable machine learning pipeline is motivated by a core limitation of penalty-based fair LTR schemes: their inability to reliably enforce fairness constraints on the predicted policies. 
A key contribution of this paper is to enable backpropagation through optimization of discontinuous OWA functions, which, in turn, has made it possible to incorporate precise group fairness measures directly into the training process of learning to rank. 
The paper showed that SOFaiR has three distinctive advantages compared to previous solutions: (1) It is able to produce rankings with high utility while also ensuring that fairness closely aligns with specified requirements; (2) it exhibits substantial efficiency improvements over other fair LTR schemes based on end-to-end optimization, delivering up to an order magnitude speedups in both training and testing; and (3) it extends naturally to fairness criteria beyond binary group treatment. These attributes may help pave the way to scalable LTR systems that are more applicable to real-world fairness requirements. These advantages also underscore the integration of constrained optimization and machine learning techniques as a promising direction for future research in fair learning to rank.

\section{Ethical Statement}
This paper was developed on commonly used, open benchmark datasets for learning to rank, and no sensitive data was used in the production of its experiments. As is common in research on fair ranking systems, protected groups were defined on the basis of attributes contained in these datasets, in order to best evaluate the performance of the algorithms. The authors' intended contribution is purely methodological, aimed at enhancing the performance of ranking systems with respect to well-established utility and fairness criteria. When considering possible unintended adverse impacts of the work, it is important to consider that the paper's methodology is generic, and can be used oppositely to its stated goals. The mechanism used to enforce fairness of group exposures in rankings can also be used to enforce arbitrary proportional exposures amongst arbitarily defined groups. Therefore, it is possible to be used in a discriminatory manner. An inherent limitation of the work, with respect to potential fairness impact, is a lack of generalization to two-sided fairness, in which fairness with respect to user utility is enforced in addition to exposure of protected groups. This stems from the fact that the machine learning methodology inherently treats each user query sample independently, thus this fairness goal is usually only pursued by post-processing methods without ML integration.

\section*{Acknowledgements}
This research is partially supported by NSF grants 2334936, 2334448, and NSF CAREER Award 2401285. 
The views and conclusions of this work are those of the authors only.

\bibliographystyle{unsrtnat}
\bibliography{sample-base}

\newpage
\appendix

\section{Additional Experimental Results}

\subsection{Multi-Group Fairness Analysis}
\label{app:multi}
\begin{figure}
    \centering
\includegraphics[width=0.45\textwidth]{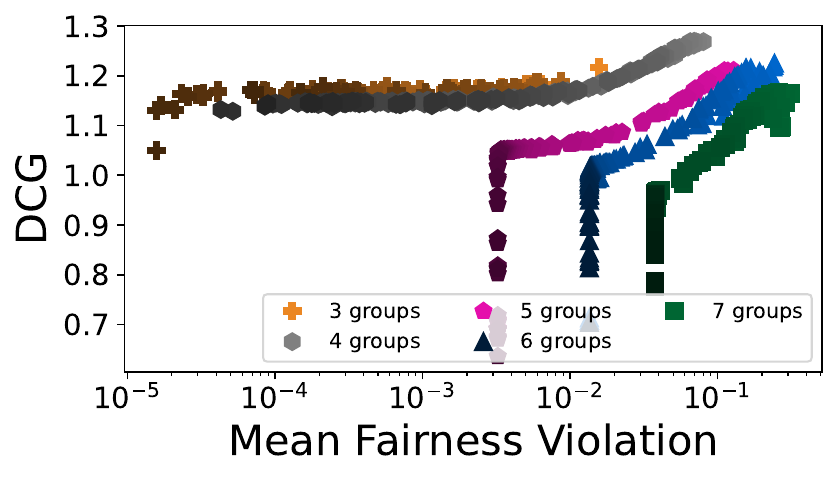}    
  \includegraphics[width=0.45\textwidth]{mslr_multi_group_utility_fairness_40_ver12.pdf}      
\includegraphics[width=0.45\textwidth]{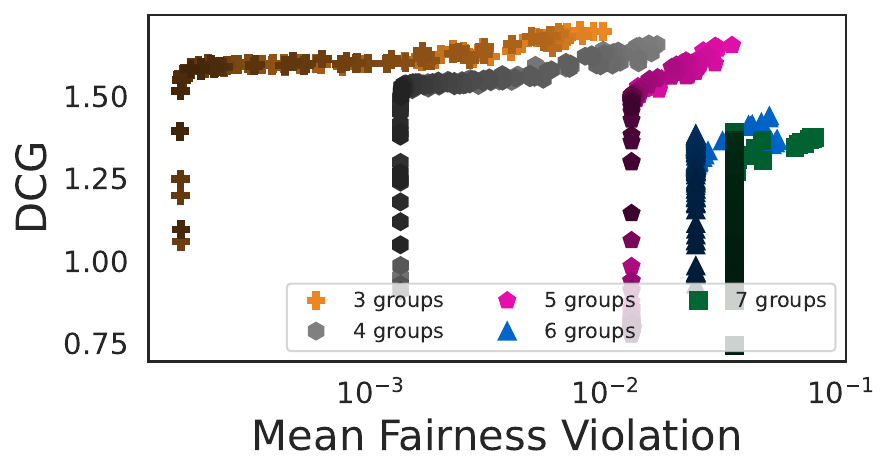}    
\includegraphics[width=0.45\textwidth]{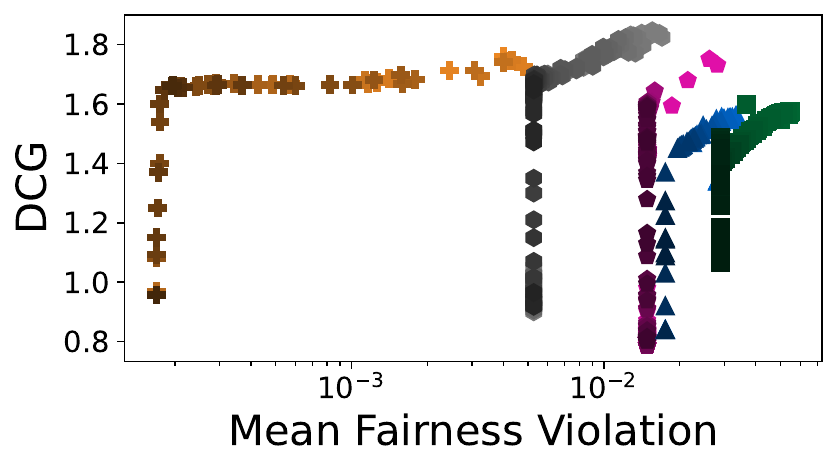}   

    \caption{Fairness-Utilities Trade-off on multi-group MSLR datasets. Top left: MSLR-20, Top right: MSLR-40, Bottom-left: MSLR-60, Bottom-right: MSLR-80 }
    \label{fig:tradeoff_mslr_multi_more}
\end{figure}

Figure \ref{fig:tradeoff_mslr_multi_more} illustrates the trade-off between utility and fairness on the other list size of MSLR-Web10k dataset (20, 60, 80 list sizes). Each data point corresponds to the performance of a single model, with fairness parameters $\lambda$ varied between 0 and 1. Models prioritizing fairness, represented by darker colored points, exhibit reduced fairness violations and lower utilities compared to those with a lower emphasis on fairness, depicted by lighter colored points. A consistent trend emerges across all datasets: as fairness parameters are relaxed, utility increases for all metrics and datasets.  Notably, saturation points in all subplots indicate that increasing the fairness weight only reduces utilities without reducing fairness violations.

\subsection{Fairness and Utility Tradeoffs Analysis}
\label{app:benchmark}

\begin{figure}
     \centering
     \begin{subfigure}[b]{0.49\linewidth}
         \centering
         \includegraphics[width=\textwidth]{mslr_binary_utility_fairness_20_ver12.pdf}
         \caption{MSLR-20}
     \end{subfigure}
     \hfill
     \begin{subfigure}[b]{0.49\linewidth}
         \centering
         \includegraphics[width=\textwidth]{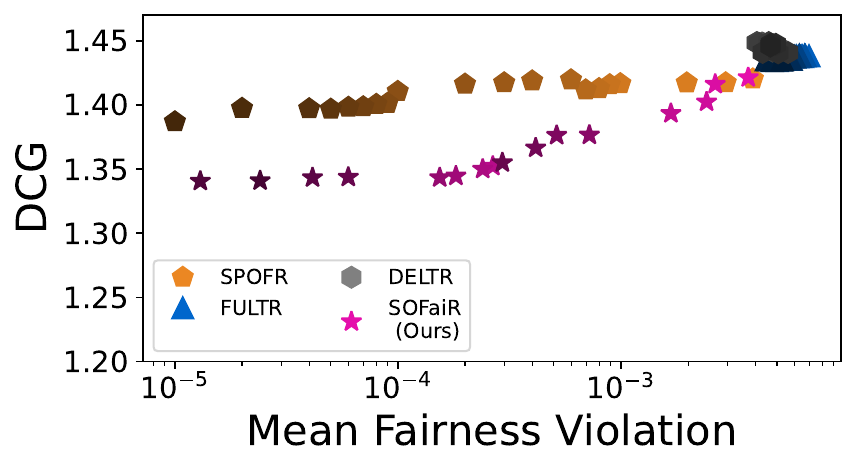}
         \caption{MSLR-40}
     \end{subfigure}
     \hfill
     
         \begin{subfigure}[b]{0.49\linewidth}
         \centering
         \includegraphics[width=\textwidth]{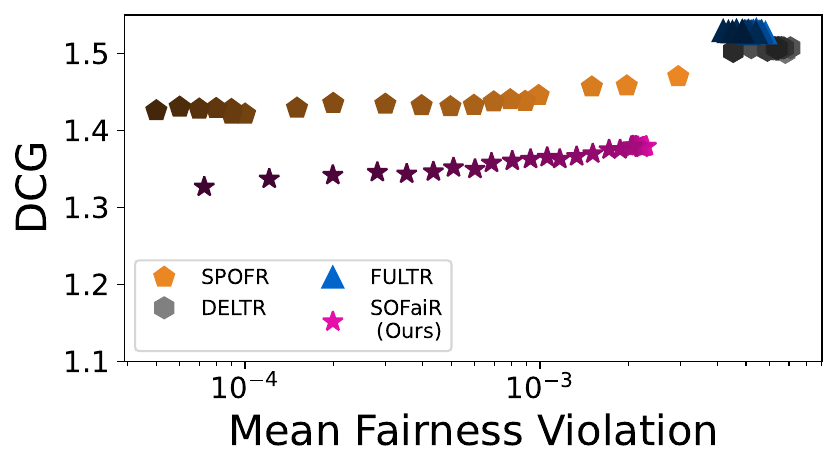}
         \caption{MSLR-60}
     \end{subfigure}
     \hfill
    \begin{subfigure}[b]{0.49\linewidth}
         \centering
         \includegraphics[width=\textwidth]{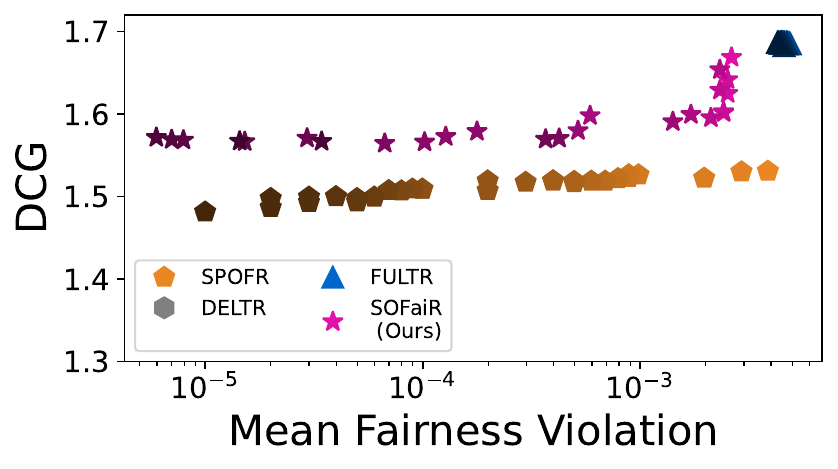}
         \caption{MSLR-80}
        
     \end{subfigure}
     \hfill
         \begin{subfigure}[b]{0.49\linewidth}
         \centering
         \includegraphics[width=\textwidth]{mslr_binary_utility_fairness_100_ver12.pdf}
         \caption{MSLR-100}
        
     \end{subfigure}
    \caption{Fairness-Utility tradeoff for MSLR datasets from all benchmark methods.}
     \label{fig:mslr_binary_full}
\end{figure}

Figure \ref{fig:mslr_binary_full} provides a comprehensive comparison of all benchmark methods in terms of fairness-utility trade-off on MSLR datasets with various list sizes. A consistent trend is observed, where fairness violations decrease at the cost of utility across all datasets for SPOFR and SOFaiR. FULTR and DELTR, while achieving high accuracy in LTR models, struggle to model fairness, evident from the constant fairness violation across different datasets.

SPOFR and SOFaiR exhibit comparable performances in terms of fairness and utility. SOFaiR, while showing slightly lower utility performance (0.05 difference for MSLR-40 and 0.1 for MSLR-60), outperforms SPOFR in larger list sizes (0.1 DCG difference in both MSLR-80 and MSLR-100).

\section{Hyper-paramaters}
\label{app:hyper}

Hyperparameters were selected as
the best-performing on average among those listed in Table \ref{app:hyper}). {Final hyperparameters for each model are as stated also in Table \ref{tab:hyperparams}, and Adam optimizer is used in the production of each result.} Asterisks (*) indicate that there is no option for a final value, as all values of each parameter are of interest in the analysis of fairness-utility tradeoff, as reported in the experimental setting Section.

For OWA optimization layers, $\bf{w}$ is set as $w_j = \frac{n-1+j}{n}$, $T=100$ during training , and $T=500$ during testing.

\begin{table}[tb]
\centering
  \caption{Hyperparameters}
\begin{tabular}{rl lllll}
\toprule
  Hyperparameter    & \multicolumn{1}{c}{Min} 
  & \multicolumn{1}{c}{Max} & 
  \multicolumn{4}{c}{Final Value} \\
  \cmidrule(r){4-7} 
  & & &  SOFaiR &SPOFR & FULTR & DELTR\\
  \midrule
  learning rate   & $1e^{-5}$ & $1e^{-1}$ &\bm{$1e^{-1}$}&$\bm{1e^{-1}}$ & $\bm{2.5e^{-4}}$ & $\bm{2.5e^{-4}}$ \\ 
  violation penalty $\lambda$   & $1e^{-5}$ & 400 & \textbf{*} & \textbf{N/A} & \textbf{*} & \textbf{*}\\ 
  allowed violation $\delta$    & 0 & 0.01 &\textbf{N/A} & \textbf{*} & \textbf{N/A} & \textbf{N/A}\\ 
  entropy regularization decay  & 0.1 & 0.5 & \textbf{N/A} & \textbf{N/A} & $\bm{0.3}$ & \textbf{N/A}\\ 
  batch size  & 64 & 512 & \textbf{256} & \textbf{256} & \textbf{256}& \textbf{256}\\
  smoothing parameter $\beta_0$ & 0.1 & 100 & \textbf{*} & \textbf{N/A} & \textbf{N/A} & \textbf{N/A} \\
 sample size & 32 & 64&\textbf{N/A} & 64 & 64 & \textbf{N/A} \\
  \bottomrule
\end{tabular}
\label{tab:hyperparams}
\end{table}

\end{document}